\documentclass[lettersize,journal]{IEEEtran}
\usepackage{amsmath,amsfonts}
\usepackage{algorithmic}
\usepackage{algorithm}
\usepackage{array}
\usepackage[caption=false,font=normalsize,labelfont=sf,textfont=sf]{subfig}
\usepackage{textcomp}
\usepackage{stfloats}
\usepackage{url}
\usepackage{verbatim}
\usepackage{graphicx}
\usepackage{cite}
\usepackage{hyperref}
\usepackage{color}
\begin{document}

% \title{Low to High Crowd Density Generalization for Crowd Counting}
\title{L2HCount: Generalizing Crowd Counting from Low to High Crowd Density via Density Simulation}

\author{Guoliang~Xu,~Jianqin~Yin,~\IEEEmembership{Member,~IEEE},~Ren~Zhang,~Yonghao~Dang, Feng~Zhou and~Bo~Yu
\thanks{Corresponding author: Jianqin Yin}
\thanks{Guoliang Xu, Jianqin Yin, Ren Zhang, Feng Zhou and Bo Yu are with the School of Intelligent Engineering and Automation, Beijing University of Posts and Telecommunications, Beijing 100876, China (e-mail:xgl@bupt.edu.cn; jqyin@bupt.edu.cn; zhangren@bupt.edu.cn; zhoufeng@bupt.edu.cn; a7858833@bupt.edu.cn;)}
\thanks{Yonghao Dang is with the School of Artificial Intelligence, Beijing University of Posts and Telecommunications, Beijing 100876, China (e-mail:dyh2018@bupt.edu.cn;)}
}

\markboth{arXiv}%
{Shell \MakeLowercase{\textit{et al.}}: A Sample Article Using IEEEtran.cls for IEEE Journals}


\maketitle

\begin{abstract}
Since COVID-19, crowd-counting tasks have gained wide applications. While supervised methods are reliable, annotation is more challenging in high-density scenes due to small head sizes and severe occlusion, whereas it’s simpler in low-density scenes. Interestingly, can we train the model in low-density scenes and generalize it to high-density scenes? Therefore, we propose a low- to high-density generalization framework (L2HCount) that learns the pattern related to high-density scenes from low-density ones, enabling it to generalize well to high-density scenes. Specifically, we first introduce a High-Density Simulation Module and a Ground-Truth Generation Module to construct fake high-density images along with their corresponding ground-truth crowd annotations respectively by image-shifting technique, effectively simulating high-density crowd patterns. However, the simulated images have two issues: image blurring and loss of low-density image characteristics. Therefore, we second propose a Head Feature Enhancement Module to extract clear features in the simulated high-density scene. Third, we propose a Dual-Density Memory Encoding Module that uses two crowd memories to learn scene-specific patterns from low- and simulated high-density scenes, respectively. Extensive experiments on four challenging datasets have shown the promising performance of L2HCount.
\end{abstract}


\section{Introduction}
\IEEEPARstart{C}{rowd} counting is an essential topic in computer vision and has been widely applied in the real world \cite{ref1,ref58}. Especially since the outbreak of COVID-19, monitoring the number of people in public places has become crucial for controlling the spread of the disease. For decades, crowd-counting methods have been divided into detection-based \cite{ref2,ref3,ref59}, density map-based \cite{ref4,ref5,ref6,ref7,ref8,ref60,ref61}, and point-based methods \cite{ref9,ref10}. Currently, the density map-based method has dominated the mainstream and achieved promising performance.

\begin{figure}[!t]
\centering
\includegraphics[width=3.5in]{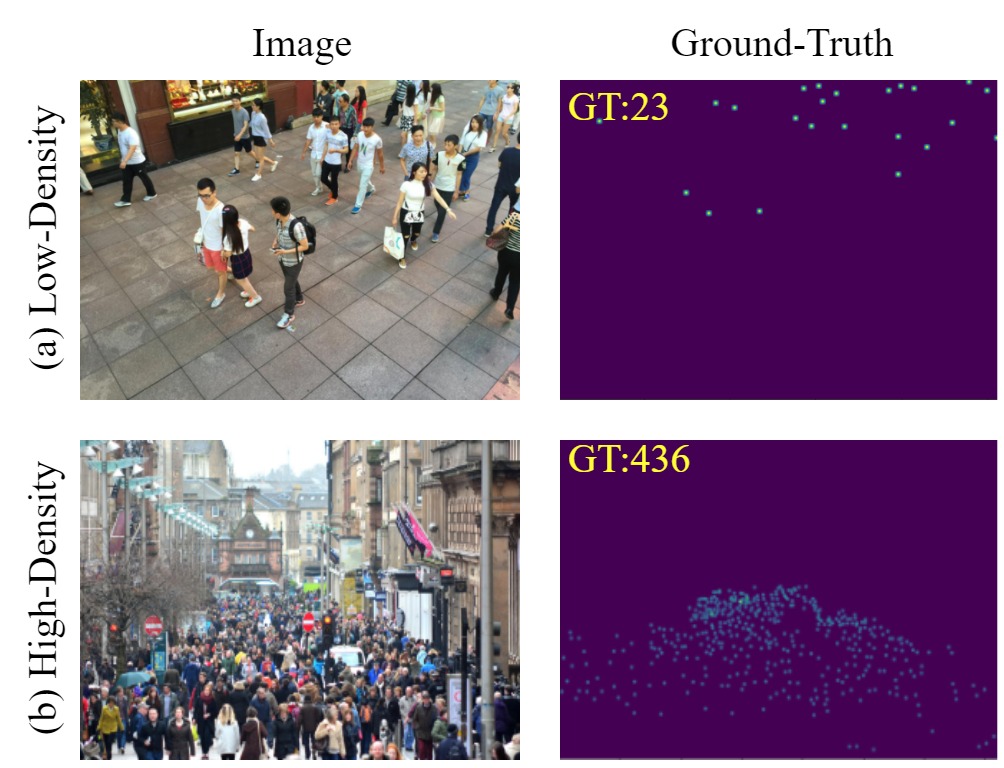}
\caption{(a) Example of image and ground-truth in the low-density scene; (b) Example of image and ground-truth in the high-density scene. GT represents the ground-truth of crowd counting in the image.}
\label{Fig1}
\end{figure}

In the crowd counting task, the annotation complexity varies across scenes with different crowd densities. Crowd data can usually be collected from different locations and environments, leading to different crowd densities, such as low- or high-crowd density. Generally, we will collect training data from specific scenes and train the model in those scenes to ensure good performance, but annotation is a very time-consuming process, especially in high-density scenes. As shown in Fig. \ref{Fig1} (b), many people have tiny head sizes in the high-density scene. When annotating, people need to enlarge the image constantly to ensure that the annotation is as accurate as possible. As shown in Fig. \ref{Fig1} (a), annotation is simpler in the low-density scene. Therefore, we have a novel idea: \textit{Can we train the model on the low-density scene and generalize it well to the high-density scene?}

Obviously, there is a noticeable domain gap between low- and high-density scenes. Domain Generalization is a common issue in crowd counting \cite{ref11,ref12,ref13,ref14}. \cite{ref11,ref12} proposed to solve the domain gap between synthetic and real data. \cite{ref13,ref14} tried to distinguish the domain-invariant and -related information and use the domain-invariant information for crowd counting in the unseen scene. Unlike them, we focus on addressing the domain gap caused by the crowd density difference.

Many methods suffer severe performance degradation from the low- to high-density scene \cite{ref13,ref14,ref15,ref16,ref17,ref18}. What exactly causes the performance degradation of these methods? Obviously, low- and high-density scenes have different characteristics. The crowd is clearly visible in the former, and the occlusion between heads is very severe in the latter. Thus, we infer that the model will learn different patterns in low- and high-density scenes, respectively, which may result in poor generalization between them. This paper aims to generalize the model well from the low- to high-density scene. \textbf{Therefore, we need to address the challenge: How can the model learn the high-density pattern from the low-density scene?}

To tackle this challenge, we propose L2HCount, a novel low- to high-crowd density generalization framework. \textbf{To simulate the high-density scene, we design a simple but effective High-Density Simualtion Module (HDSM), enabling simulation of the characteristics in the high-density scene: dense crowd distribution and head occlusion.} Specifically, we shift the original image of the low-density scene by some pixels and then overlay it with the original image by a ratio to construct the fake high-density image. Additionally, we also propose a Ground-Truth Generalization Module (GTGM) to automatically generate the corresponding ground-truth crowd annotations of the simulated high-density images for supervised learning.

However, the simulated high-density images introduce two new problems: image blurring and loss of low-density image characteristics. For the first problem, the head area overlaps with the background in the simulated high-density image, making it difficult for the model to distinguish between the features of the head and background. Therefore, to solve this problem, we propose the Head Feature Enhancement Module (HFEM). Specifically, since the clarity of low-density images, we can extract distinct features from them. Then, similar to HDSM, we also shift and overlay these features to construct features of simulated high-density images, using the constructed features as supervision to guide the model for learning the discriminative features of the head and background in the high-density images. For the second problem, to make our model learn the patterns of low- and high-density scenes simultaneously, we also propose the Dual-Density Memory Encoding Module (DDMEM), which uses the Low-Density Crowd Memory (LDCM) and High-Density Crowd Memory (HDCM) to learn the scene-specific patterns of low- and high-density scenes, respectively.

To sum up, this paper addresses the challenge of generalizing the model from the low- to high-density scene. The main contributions are as follows.
\begin{itemize}
  \item We propose L2HCount, a framework for training the model in the low-density scene while enabling it to generalize effectively to the high-density scene.
  \item We design the HDSM to simulate the high-density scene using the low-density scene and the GTGM to automatically generate the GT of the simulated high-density scene. We also propose the HFEM and DDMEM. The former guides the model to distinguish the head and background features, and the latter allows the model to simultaneously learn the patterns of low- and high-density scenes.
  \item Compared with state-of-the-art methods, our approach achieves promising counting performance on popular benchmarks.
\end{itemize}

\section{Related works}
\subsection{Crowd Counting}
\textbf{Detection-based methods:} These methods aim to detect or segment each human body and further count the number of crowds \cite{ref19,ref20,ref21}. \cite{ref19} combined a MID (Mosaic Image Difference)-based foreground segmentation method and a HOG (Histograms of Oriented Gradients)-based head-shoulder detection method to estimate the people count in surveillance scenes. Lin et al. \cite{ref20} proposed a shape-based, hierarchical part-template matching method to achieve human detection and segmentation for crowd counting. \cite{ref21} introduced a trajectory clustering method to count the number of moving crowds in a scene. In crowd-counting tasks, occlusion is a serious challenge. It is difficult to handle this challenge using detection-based methods.

\textbf{Point-based methods:} These methods aim to predict the center point of each head for crowd counting \cite{ref15, ref16, ref10, ref22, ref23, ref24}. It is natural to use point annotations to supervise networks directly. However, this process must carefully design the loss to ensure a good correspondence between predicted points and ground-truth. Bayesian+ \cite{ref15} proposed a novel Bayesian loss to construct a density contribution probability model from point annotation, which is more robust to occlusions, perspective effects, etc. DM-Count \cite{ref16} used Optimal Transport (OT) to measure the similarity between the predicted result and the ground-truth, which improves the model's generalization. Like \cite{ref16}, UOT \cite{ref22} adopted the unbalanced OT distance, which remains stable under spatial perturbations, to measure the difference between the predicted results and point annotations. \cite{ref23} proved that pixel-wise L2 loss and Bayesian loss are suboptimal solutions. Therefore, they further proposed a perspective-guided transport cost function for crowd counting and localization. To further improve individual localization, P2PNet \cite{ref10} directly predicted a set of point proposals to represent heads in an image, aligning closely with human annotations. Further, APGCC \cite{ref24} improved the point-based crowd counting. They proposed auxiliary point guidance to offer clear guidance for point proposal selection and optimization, addressing the core challenge of matching uncertainty. 

\textbf{Density map-based methods:} These methods aim to predict a density map and further compute the sum of all pixel values in the density map to obtain the crowd count. Before training, we need to convert an image with labeled head positions to the crowd density map by the Gaussian kernel as ground truth. Existing methods typically use Convolutional Neural Network (CNN) \cite{ref25,ref26,ref27,ref28,ref29,ref30,ref31}, Transformer \cite{ref5,ref7,ref32,ref33,ref34}, and Mamba to extract image features for density map prediction.

Due to the influence of perspective, the heads have different sizes in the same image. Therefore, \cite{ref25,ref26} used the multi-column convolutional neural network with receptive fields of different sizes to extract the multi-scale features for crowd counting. SANet \cite{ref27} proposed a novel encoder-decoder network with CNN. The encoder used the scale aggregation modules to extract multi-scale features, and the decoder used transposed convolutions to generate high-resolution density maps. SASNet \cite{ref31} proposed a scale-adaptive selection network that automatically learns the internal correspondence in different scales. In addition to head size factors, other factors, such as background clutters, distribution of people, etc., can also affect the counting performance of the model. CRNet \cite{ref28} can refine predicted density maps progressively based on hierarchical multi-level density priors, which improve the model's robustness. HA-CCN \cite{ref29} used attention mechanisms at various levels to selectively refine the features, which improved the model's performance in highly congested scenes. PaDNet \cite{ref30} designed a density-aware network to count people in varying-density crowds.

Compared with CNN-based methods, Transformer-based methods \cite{ref5,ref7,ref32,ref33,ref34} have the ability of global modeling to exhibit strong features. MAN \cite{ref32} incorporated global attention from the vanilla transformer, learnable local attention, and instance attention into a counting model. AGCCM \cite{ref33} designed two kinds of bidirectional transformers to decouple the global attention, including row and column attention, which reduced the computational complexity. \cite{ref34} proposed to build the top-down visual perception mechanism with Transformen in crowd counting. Gramformer \cite{ref7} proposed a graph-modulated transformer to enhance the network by adjusting the attention and input node features, respectively, which can solve the problem of the homogenized solution. Due to the respective advantages of CNN and transformer, CTASNet \cite{ref5} proposed a CNN and transformer adaptive selection network. This network can adaptively select the appropriate counting branch (CNN-based and Transformer-based branches are responsible for low-density and high-density regions, respectively) for different density regions.

The transformer-based method has an obvious problem: the quadratic complexity of self-attention w.r.t the number of tokens imposes a substantial computational cost. Therefore, Mamba \cite{ref36,ref37} integrated the selective structured state space models into a simplified end-to-end network without the attention mechanism and MLP. Mamba has the advantages of fast inference and linear complexity. Subsequently, VMamba \cite{ref38} used Mamba to build a vision backbone with linear time complexity. \cite{ref35} applied VMamba to crowd counting and proposed a VMamba Crowd Counting (VMambaCC) that inherits the merits of VMamba, namely, global modeling for images and low computational cost.

Density map-based methods have become increasingly popular due to their promising counting accuracy. Our method also falls into this category.

\subsection{Domain Adaptation for Crowd Counting}
Crowd counting has significant practical value. However, real-world scenes may not always be consistent with training scenes. Therefore, cross-domain methods are needed to maintain the model's performance in unseen scenes.

Domain Adaptation (DA) \cite{ref39,ref40,ref41,ref42,ref43} aims to train the model using data from the source domain and then fine-tune it using a small amount of data from the target domain. \cite{ref39} proposed an innovative framework called explicit invariant feature induced cross-domain knowledge transformation framework to address the inconsistency of domain-invariant features across different domains. This method aims to extract the invariant features from source and target domains. Due to the higher annotation cost, some methods use synthetic crowd data as the source domain and transfer knowledge to real data \cite{ref40,ref41,ref42,ref43}. DACC \cite{ref40} proposed high-quality image translation and density map reconstruction modules. The former focused on translating synthetic data to realistic data, and the latter aimed to generate the pseudo labels for training on real data. \cite{ref41} proposed a large-scale synthetic dataset. They pre-trained the model on this dataset and fine-tuned it on real datasets. BLA \cite{ref42} proposed task-driven data alignment and fine-grained feature alignment to reduce the gap between synthetic and real data. \cite{ref43} proposed two-stage methods in which synthetic and real data are input to model simultaneously in the entire process. During the first stage, the shared backbone is used to extract features of both data, which can reduce the domain gap of both data. During the second stage, they used the uncertainty of the density map to generate pseudo labels for further model fine-tuning.

However, DA still requires data from target scenes to optimize the network. Such data is difficult to obtain in unseen target scenes.

\subsection{Domain Generalization for Crowd Counting}

Unlike DA, Domain Generalization (DG) aims to train the model using only data from the source domain and directly perform inference on the target domain, which is more aligned with real-world applications.

In DG-based methods, the key is to extract domain-invariant features from the source domain. \cite{ref13} dynamically divided the source domain into a series of sub-domains. Then, they used the meta-learning framework for domain generalization. In each iteration, they sampled one image from each sub-domain, selecting one image as the meta-testing image and the remaining ones as meta-training images. To distinguish the domain-invariant and -specific features, they designed the domain-invariant and -specific crowd memory modules to learn different feature representations, respectively. Finally, the domain-invariant crowd memory module was used to encode the image feature for crowd counting. Inspired by \cite{ref13}, MPCount \cite{ref14} designed an Attention Memory Bank (AMB) and a Content Error Mask (CEM) for DG. Specifically, MPCount used a pair of images, consisting of an original image and its data-augmented version, as input. Then, it employed the CEM to eliminate domain-specific content from image features. Then, the remaining content (domain-invariant content) was input to the AMB to reconstruct the domain-invariant features for crowd counting.

Crowd density is also an important factor causing domain gap, which has been overlooked by other methods \cite{ref13,ref14}. We propose a novel idea: Can we train the model in low-density scenes and generalize it to high-density scenes? To train a model that performs reliably in the target scene, data collection and annotation from that scene are typically required. However, this process poses significant challenges in real-world applications. First, obtaining data from the target scene in advance is often impractical. Second, even if data is available, annotation remains a challenging problem, especially in high-density scenes. To address these issues, we propose L2HCount, a novel low- to high-crowd density generalization framework.

\section{Proposed Method}

\begin{figure*}[!t]
\centering
\includegraphics[width=7.2in]{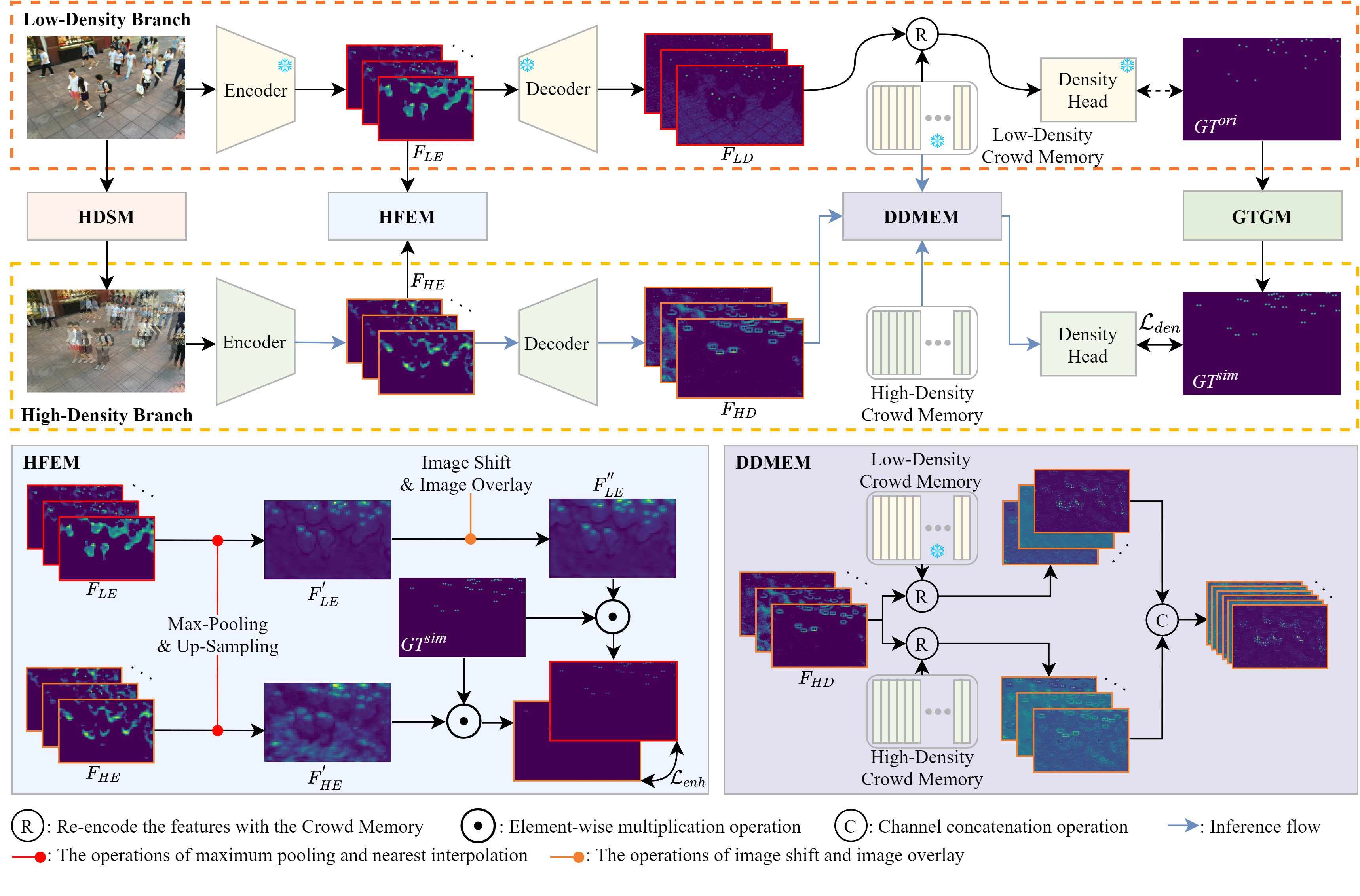}
\caption{The proposed L2HCount is used to realize the crowd counting generalization from the low- to high-density scene. The L2HCount mainly contains high- and low-density branches, the High-Density Simulation Module (HDSM), the Head Feature
Enhancement Module (HFEM), the Dual-Density Memory Encoding Module (DDMEM), and the Ground-Truth Generation Module
(GTGM). $GT^{ori}$ and $GT^{sim}$ represent the ground-truth of low- and high-density images. It is worth noting that we only use the high-density branch and DDMEM for inference, as shown in blue arrows.}
\label{Fig2}
\end{figure*}

Fig. \ref{Fig2} presents an overview of the proposed method, L2HCount, which consists of two branches (low- and high-density branches) and four modules (HDSM, HFEM, DDMEM, and GTGM). The low-density branch takes low-density images as input and employs the Low-Density Crowd Memory (LDCM) to learn the pattern of low-density scenes. The high-density branch takes simulated high-density images as input and utilizes the High-Density Crowd Memory (HDCM) to learn the pattern of high-density scenes. To simultaneously handle both low- and high-density situations, we re-encode the density features using LDCM and HDCM, respectively. Then, the re-encoded features are concatenated along the channel dimension to predict the density map for crowd counting. It is worth noting that only the high-density branch and DDMEM are used for model inference.

The following sections provide a detailed introduction to the four modules mentioned earlier.

\subsection{High-Density Simulation Module}
Apparent differences exist between low- and high-density scenes, such as crowd density and head occlusion. Therefore, a key problem is how to simulate high-density scenes using low-density scenes. We propose a natural approach: shifting original low-density images and overlapping them with the original images to simulate high-density images. This approach effectively mitigates two major differences. First, the simulated image’s crowd count is doubled, increasing the overall crowd density. Second, the simulated image contains more head occlusions, making it more consistent with the characteristics of high-density scenes.

Specifically, as shown in Eq. \ref{eq1}, we can easily achieve image shift and overlay using zero-padding operations. ${P_L}( \cdot , S )$ and ${P_R}( \cdot , S )$, respectively, represent zero-padding operations on the left and right sides of the image. $S$ controls the amount of padding in the image. $\lambda$ represents the ratio of image fusion. In this paper, we set $\lambda  = 0.5$. ${I^{ori}} \in {R^{H \times W \times 3}}$ and ${I^{sim}} \in {R^{H \times (W+S) \times 3}}$ represent the original image and simulated image, respectively.

\begin{equation}
\label{eq1}
{I^{sim}} = \lambda  \times {P_L}({I^{ori}},S) + (1 - \lambda ){P_R}({I^{ori}},S)
\end{equation}

Although HDSM is used to simulate high-density scenes, this module introduces two new problems: image blurring and the loss of low-density image characteristics. Therefore, we propose HFEM and DDMEM to address these challenges, respectively.

\subsection{Head Feature Enhancement Module}
In simulated images, the background and head regions overlap, making it difficult for the encoder of the high-density branch to extract clear head features for crowd counting. In contrast, the input image of the low-density branch is clear, allowing its encoder to extract distinct head features. A natural question arises: Can the features of the low-density branch be used to guide the feature learning process in the high-density branch?

In detail, as illustrated in Fig. \ref{Fig2}, we define the output features of the encoder in low- and high-density branches as $F_{LE}\in {R^{c_1 \times (w/16) \times (h/16)}}$ and $F_{HE} \in {R^{c_1 \times (w/16) \times (h/16)}}$, respectively. We find that $F_{LE}$ generates more significant activation at the head locations. Therefore, we perform max-pooling on the channel dimensions of $F_{LE}$ and $F_{HE}$. Since the simulated image is obtained by image shift and overlap on the original image, we first up-sample the feature by nearest interpolation to the same size as the original image. Then, we achieve the feature shift and overlap on $F_{LE}^{'}$ to obtain $F_{LE}^{''}$. To enable the encoder of the high-density branch to extract clear head features, we formulate the MSE loss between the head locations of $F_{HE}^{'}$ and $F_{LE}^{''}$, as shown in Eq. \ref{eq2}.

\begin{equation}
\label{eq2}
{\mathcal{L}_{enh}} = {\rm{MSE}}(F_{HE}^{'} \odot G{T^{sim}},F_{LE}^{''} \odot G{T^{sim}})
\end{equation}

\subsection{Dual-Density Memory Encoding Module}
Due to the limitation of HDCM, the simulated high-density image loses the characteristics of low-density images, making the model unable to simultaneously learn both low- and high-density patterns from the simulated image. However, this limitation affects the model's performance in real-world applications. This is because, due to the perspective effect, low-density and high-density regions may coexist within the same image, even in high-density scenes. Therefore, we propose DDMEM, which enables the model to handle both low- and high-density situations simultaneously.

Specifically, we design Low-Density Crowd Memory (LDCM) and High-Density Crowd Memory (HDCM) to learn low- and high-density patterns. The LDCM ${V_{LDCM}} \in {R^{c_2 \times l}}$ and HDCM ${V_{HDCM}} \in {R^{c_2 \times l}}$
consist of $l$ memory vectors of dimension $c_2$, respectively. First, we obtain the flattened feature ${F_{HD}} \in {R^{{c_2} \times hw}}$ of the decoder in the high-density branch. As shown in Eq. \ref{eq3}, we compute the attention scores ${A_{LD}} \in {R^{hw \times {l}}}$ between the feature ${F_{HD}}$ and the LDCM ${V_{LDCM}}$. In the same way, we also compute the attention scores ${A_{HD}} \in {R^{hw \times {l}}}$ between the feature ${F_{HD}}$ and the HDCM ${V_{HDCM}}$. To reconstruct the crowd density-specific features, we use $A_{LD}$ and $V_{LDCM}$ to reconstruct low-density-specific feature ${{\tilde F}_{LD}} \in {R^{{c_2} \times hw}}$ and use $A_{HD}$ and $V_{HDCM}$ to reconstruct high-density-specific feature ${{\tilde F}_{HD}} \in {R^{{c_2} \times hw}}$, respectively. ${{\tilde F}_{LD}}$ is more suitable for estimating the crowd count of the low-density region, while ${{\tilde F}_{HD}}$ is more suitable for estimating the crowd count of the high-density region. Therefore, as illustrated in Eq. \ref{eq4}, we fuse the ${{\tilde F}_{LD}}$ and ${{\tilde F}_{HD}}$ by feature concatenation to obtain the more abundant feature ${\tilde F}$ for crowd counting.

\begin{equation}
\label{eq3}
A({F_{HD}},V) = {\rm{Softmax}}(\frac{{{F_{HD}^T}V}}{{\sqrt c_2 }})
\end{equation}

\begin{equation}
\label{eq4}
\tilde F = {\rm{Concat}}({V_{LDCM}{{A_{LD}}^T}},{V_{HDCM}{{A_{HD}}^T}})
\end{equation}

\subsection{Ground-Truth Generation Module}
In this paper, we hope to avoid manually annotating the simulated high-density images. Therefore, we propose GTGM, which automatically marks the position of each human head in simulated images using annotations from low-density images.

Specifically, as shown in Eq. \ref{eq5}, we also use the zero-padding operations to achieve the shift and overlay of $G{T^{ori}}$ for obtaining the $G{T^{sim}}$. Where $G{T^{ori}} \in {R^{H \times W \times 3}}$ is the $GT$ of the low-density image, and $G{T^{sim}} \in {R^{H \times (W+S) \times 3}}$ is the generated $GT$ of the simulated high-density image.

\begin{equation}
\label{eq5}
G{T^{sim}} = {P_L}(G{T^{ori}},S) + {P_R}(G{T^{ori}},S)
\end{equation}

\subsection{Loss Function}

The overall training loss combines density loss $\mathcal{L}_{den}$, head feature enhancement loss $\mathcal{L}_{enh}$, patch-wise classification loss $\mathcal{L}_{cls}$, and attention consistency loss $\mathcal{L}_{con}$. The third and fourth terms in Eq. \ref{eq6} are not the ideas of this paper, which were proposed in \cite{ref14}. Therefore, we do not introduce their specific details in our paper.

\begin{equation}
\label{eq6}
\mathcal{L} = {\theta _{den}}{\mathcal{L}_{den}} + {\theta _{enh}}{\mathcal{L}_{enh}} + {\theta _{cls}}{\mathcal{L}_{cls}} + {\theta _{con}}{\mathcal{L}_{con}}
\end{equation}

Where $\theta _{den}$, $\theta _{enh}$, $\theta _{cls}$, and $\theta _{con}$ are weighting coefficients to balance different loss terms.

\section{Experiments}

In this section, we conduct various experiments to evaluate the proposed L2HCount. First, we introduce the experimental setup in detail, including the network architecture, training details, evaluation metrics, and datasets. Then, we compare our method with state-of-the-art approaches to demonstrate its superior performance. Finally, we perform a series of ablation studies to validate the effectiveness of each proposed module.

\subsection{Experimental Setups}

\textit{1) Network Architecture:} Our proposed L2HCount consists of a low-density branch and a high-density branch. This paper aims to generalize the model from the low-density scene to the high-density scene rather than the innovation of detailed network structure. Therefore, we design both branches by referring to \cite{ref14}. Both branches have a similar structure but do not share parameters. We adopt VGG16-BN \cite{ref44} as the encoder to extract image features in our model. The decoder consists of six $3\times3$ convolution layers followed by a $1\times1$ convolution layer. Each layer consists of a convolution operation, batch normalization, and ReLU function. The density head consists of a $1\times1$ convolution operation followed by a ReLU activation function. LDCM and HDCM consist of $l$ memory vectors, each with a dimension of $c_2$. The memory size $l$ is set to 1024, and the dimension $c_2$ is 256.

\textit{2) Training Details:} The existing methods have achieved promising performance in low-density scenes, so we do not re-design a new method as our low-density branch. We directly load the network structure and parameters of MPCount \cite{ref14} as the low-density branch. We also freeze the parameters of the low-density branch during the training phase. In addition, we use AdamW \cite{ref45} as the optimizer and OneCycleLR \cite{ref46} as the learning rate scheduler with a maximum learning rate of 1e-4 and a maximum epoch of 180 to optimize the other parameters. The weighting coefficients $\theta _{den}$, $\theta _{enh}$, $\theta _{cls}$, and $\theta _{con}$ are set to 1, 1, 10, and 10.

Like \cite{ref14}, we also use data augmentation, including three types of photometric transformations: color jittering, Gaussian blurring, and sharpening. We also randomly crop image patches with the size of $w\times h$ ($320\times320$) and randomly use horizontal flipping to both the low- and simulated high-density images.

\textit{3) Evaluation Metrics:} We evaluate our method using the Mean Absolute Error (MAE) and  Mean Squared Error (MSE), defined as follows:

\begin{equation}
\label{eq7}
MAE = \frac{1}{N}\sum\limits_{i = 1}^N {|{c_i} - {{\hat c}_i}|}
\end{equation}

\begin{equation}
\label{eq8}
MSE = \sqrt {\frac{1}{N}\sum\limits_{i = 1}^N {{{({c_i} - {{\hat c}_i})}^2}} } 
\end{equation}

Where $N$ is the number of testing images, $c_i$ is the crowd count of the $i-$th image predicated by the network, and ${{{\hat c}_i}}$ is the GT of the $i-$th image.

\begin{figure}[!t]
\centering
\includegraphics[width=3.5in]{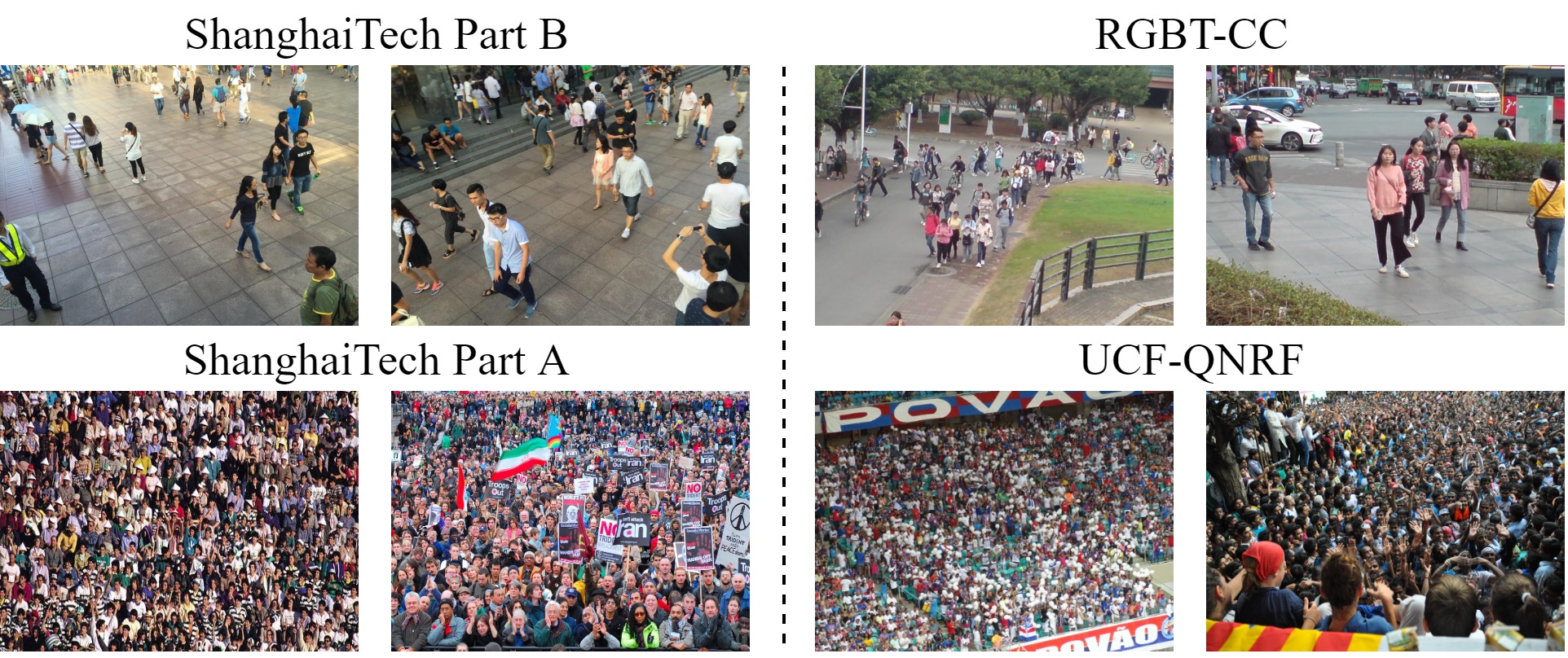}
\caption{Some samples of images from ShanghaiTech Part A$\&$B, RGBT-CC, and UCF-QNRF datasets.}
\label{Fig9}
\end{figure}

\textit{4) Datasets:} We aim to verify the model's generalization ability from low- to high-density scenes. Therefore, our proposed model would train on the low-density dataset and then directly test on the high-density dataset. We select four mainstream crowd-counting datasets: ShanghaiTech Part A $\&$ B, RGBT-CC, and UCF-QNRF. As shown in Fig. \ref{Fig9}, ShanghaiTech Part B and RGBT-CC have a low-density crowd distribution, while ShanghaiTech Part A and UCF-QNRF have a high-density crowd distribution.

\begin{itemize}
  \item ShanghaiTech Part B (B) \cite{ref25} includes 716 images, 400 of which are used for training and 316 for testing. Part B is taken from the busy streets of Shanghai's metropolitan areas.
  \item RGBT-CC (R) \cite{ref57} is a large-scale RGBT crowd-counting benchmark. It consists of 1013 pairs of images captured in the light and 1017 pairs of images captured in the darkness, including various scenes (e.g., malls, streets, playgrounds, train stations, metro stations, etc.). In this paper, we only use the RGB images captured in the light.
  \item ShanghaiTech Part A (A) \cite{ref25} includes 482 images, 300 of which are used for training and the remaining for testing. These images are randomly collected from the Internet. 
  \item UCF-QNRF (Q) \cite{ref47} consists of 1201 training images and 334 testing images. It is a challenging dataset that is captured from different locations, viewpoints, and lighting conditions.
\end{itemize}

\subsection{Comparison With State-of-the-Art (B$\to$A/Q)}

In this section, we compare our L2HCount with state-of-the-art methods on different benchmarks, as shown in Table \ref{tab:table1}. The selected methods consist of fully supervised methods, domain adaptation-based methods, and domain generalization-based methods, as follows:

\begin{itemize}
  \item \textit{Fully supervised methods:} BL \cite{ref15}, DMCount \cite{ref16}, SASNet \cite{ref31}, ChfL \cite{ref48}, and MAN \cite{ref32}.
  \item \textit{Domain adaptation-based methods:} RBT \cite{ref49}, C$^2$MoT \cite{ref50}, FGFD \cite{ref51}, and FSIM \cite{ref52}.
  \item \textit{Domain generalization-based methods:} IBN \cite{ref53}, SW \cite{ref54}, ISW \cite{ref55}, DG-MAN \cite{ref56}, DCCUS \cite{ref13}, and MPCount \cite{ref14}.
\end{itemize}

\begin{table*}[!t]
\caption{Comparison with state-of-the-art methods (Training on the B dataset and testing on A and Q datasets). Bold numbers represent the best performance, and underlined numbers represent the second-best performance. B $\to$ A and B $\to$ Q represent training the model on the training set of B and then directly verifying the model's performance on the testing set of A or Q. A $\to$ A and Q $\to$ Q represent training the model on the training sets of A and Q respectively, and then directly verifying the model's performance on the testing sets of A and Q datasets.
\label{tab:table1}}
\setlength\tabcolsep{11pt}
\centering
\begin{tabular}{ccccccccccc}
\hline
\multicolumn{3}{c}{Source$\to${}Target} & \multicolumn{2}{c}{B$\to$A} & \multicolumn{2}{c}{B$\to${}Q} & \multicolumn{2}{c}{A$\to${}A} & \multicolumn{2}{c}{Q$\to${}Q} \\ \hline
Method                & DA         & DG         & MAE               & MSE               & MAE               & MSE         & MAE               & MSE               & MAE               & MSE               \\ \hline
BL \cite{ref15}                    & $\times$          & $\times$          & 138.1             & 228.1             & 226.4             & 411.0          & 62.8             & 101.8             & 88.7             & 154.8             \\
DMCount \cite{ref16}               & $\times$          & $\times$          & 143.9             & 239.6             & 203.0             & 386.1          & 59.7             & 95.7             & 85.6             & 148.3             \\
SASNet \cite{ref31}                & $\times$          & $\times$          & 132.4             & 225.6             & 273.5             & 481.3          & \textbf{53.59}             & \textbf{88.38}             & 85.2             & 147.3             \\
ChfL \cite{ref48}                  & $\times$          & $\times$          & 121.3             & 200.8             & 197.1             & 357.9          & 57.5             &  94.3             &  \underline{80.3}             & \underline{137.6}             \\
MAN \cite{ref32}                   & $\times$          & $\times$          & 133.6             & 255.6             & 209.4             & 378.8          & \underline{56.8}             & \underline{90.3}             & \textbf{77.3}             & \textbf{131.5}             \\ \hline
RBT \cite{ref49}                   & $\checkmark$          & $\times$          & 112.2             & 218.2             & 211.3             & 381.9          & -             & -             & -             & -             \\
C$^2$MoT \cite{ref50}              & $\checkmark$          & $\times$          & 120.7             & 192.0             & 198.9             & 368.0          & -             & -             & -             & -             \\
FGFD \cite{ref51}                  & $\checkmark$          & $\times$          & 123.5             & 210.7             & 209.7             & 384.7          & -             & -             & -             & -             \\
FSIM \cite{ref52}                  & $\checkmark$          & $\times$          & 120.3             & 202.6             & 194.9             & 324.5          & -             & -             & -             & -             \\ \hline
IBN \cite{ref53}                   & $\times$          & $\checkmark$          & 125.9             & 202.3             & 183.5             & 317.4          & -             & -             & -             & -             \\
SW \cite{ref54}                    & $\times$          & $\checkmark$          & 126.7             & 193.8             & 200.7             & 333.2          & -             & -             & -             & -             \\
ISW \cite{ref55}                   & $\times$          & $\checkmark$          & 156.2             & 291.5             & 263.1             & 442.2          & -             & -             & -             & -             \\
DG-MAN \cite{ref56}                & $\times$          & $\checkmark$          & 130.7             & 225.1             & 182.4             & 325.8          & -             & -             & -             & -             \\
DCCUS \cite{ref13}                 & $\times$          & $\checkmark$          & 121.8             & 203.1             & 179.1             & 316.2          & -             & -             & -             & -             \\
MPCount \cite{ref14}               & $\times$          & $\checkmark$          & \underline{99.6}              & \underline{182.9}             & \underline{165.6}             & \underline{290.4}          & -             & -             & -             & -             \\
L2HCount(Ours)         & $\times$          & $\checkmark$          & \textbf{97.6}              & \textbf{166.9}             & \textbf{159.4}             & \textbf{267.7}          & -             & -             & -             & -             \\ \hline
\end{tabular}
\end{table*}

\textit{Comparison with fully supervised methods:} As shown in Table \ref{tab:table1}, we compare our method with the fully supervised method under two training settings. Training setting 1: The source domain and target domain are different datasets. Training setting 2: The source and target domains are the same dataset.

Compared with training setting 2, the fully supervised methods show a noticeable performance degradation under training setting 1. The reason is that a domain shift problem is caused by crowd density between the source domain and the target domain under training setting 1, which is a challenge of the fully supervised methods.

Under training setting 1, the performance of L2HCount outperforms all fully supervised methods. When tested on the A dataset, L2HCount achieves 19.54$\%$ (97.6 vs. 121.3) MAE and 16.88$\%$ (166.9 vs. 200.8) MSE improvement compared with ChfL. Compared with DMCount, L2HCount achieves 32.18$\%$ (97.6 vs. 143.9) MAE and 30.34$\%$ (166.9 vs. 239.6) MSE improvement, respectively. When tested on the Q dataset, L2HCount shows a 19.13$\%$ (159.4 vs. 197.1) improvement in MAE and a 25.20$\%$ (267.7 vs. 357.9) improvement in MSE compared with ChfL. Compared with DMCount, L2HCount shows a 21.48$\%$ (159.4 vs. 203.0) improvement in MAE and a 30.67$\%$ (267.7 vs. 386.1) improvement in MSE, respectively. These results indicate that our method effectively alleviates the domain shift caused by crowd density, allowing our model to maintain its performance in the high-density scene.

\textit{Comparison with domain adaptation-based methods:} In addition to training the model with source domain data, domain adaptation-based methods typically require fine-tuning it with a few of the target domain samples. L2HCount also outperforms all domain adaptation-based methods. When tested on the A dataset, L2HCount shows 18.87$\%$ (97.6 vs. 120.3) improvement in MAE and 17.62$\%$ (166.9 vs. 202.6) improvement in MSE compared to FSIM. Against RBT, it improves by 13.01$\%$ (97.6 vs. 112.2) in MAE and 23.51$\%$ (166.9 vs. 218.2) in MSE. On the Q dataset, L2HCount achieves 18.21$\%$ (159.4 vs. 194.9) improvement in MAE and 17.50$\%$ (267.7 vs. 324.5) improvement in MSE compared to FSIM. Compared to RBT, L2HCount shows 24.56$\%$ (159.4 vs. 211.3) improvement in MAE and 29.90$\%$ (267.7 vs. 381.9) improvement in MSE. These results demonstrate that our method effectively alleviates domain shift issues caused by crowd density. We only need low-density images from the source domain to train our model, avoiding the need for fine-tuning with target domain data. Therefore, when applied to high-density scenes, our method eliminates the additional labor costs associated with data annotation.

\textit{Comparison with domain generalization-based methods:} We compared our method with domain generalization-based approaches, and it still performs best. All methods are trained on low-density datasets (B) and directly tested on high-density datasets (A and Q). On the A dataset, L2HCount shows 2.01$\%$ (97.6 vs. 99.6)  improvement in MAE and 8.75$\%$ (166.9 vs. 182.9) improvement in MSE compared to MPCount, and outperforms DCCUS by 19.87$\%$ (97.6 vs. 121.8) in MAE and 17.82$\%$ (166.9 vs. 203.1) in MSE. On the Q dataset, L2HCount achieves a 3.74$\%$ (159.4 vs. 165.6) improvement in MAE and a 7.82$\%$ (267.7 vs. 290.4) improvement in MSE compared to MPCount. Against DCCUS, L2HCount achieves 11.00$\%$ (159.4 vs. 179.1) improvement in MAE and 15.34$\%$ (267.7 vs. 316.2) improvement in MSE.

We analyze two key factors that contribute to the promising performance of L2HCount. First, we generate simulated high-density images from low-density images using HDSM, enabling our model to adapt to the challenges of high-density scenes, such as head occlusion and crowd distribution. Second, we design LDCM and HDCM to learn the patterns of low- and high-density images, allowing our model to handle both situations simultaneously.

\begin{figure}[!t]
\centering
\includegraphics[width=3.5in]{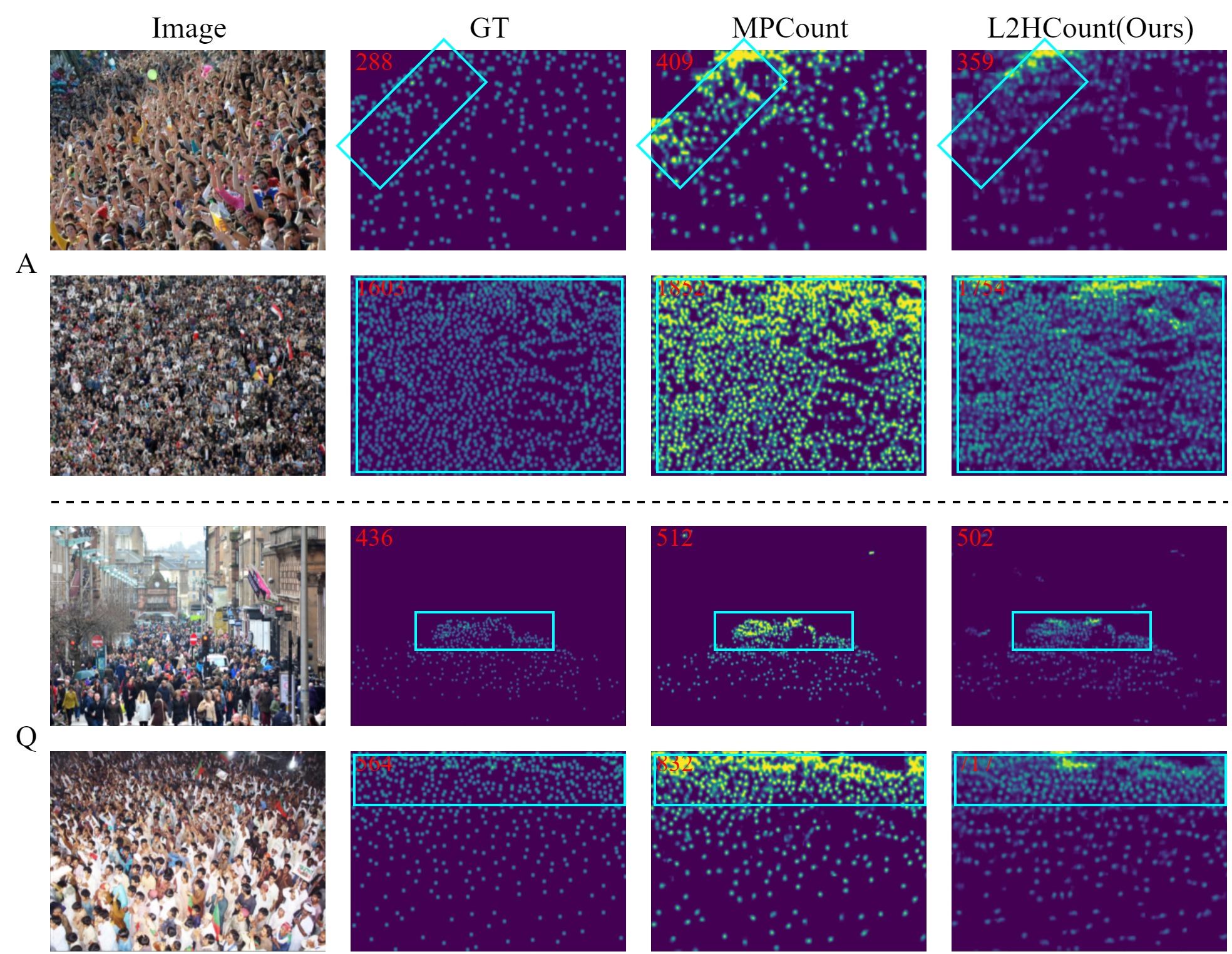}
\caption{Visualization of the predicted results. We show the predicted results of MPCount and L2HCount on the A and Q datasets. L2HCount predicts more reasonable results in the rectangle regions.}
\label{Fig3}
\end{figure}

Fig. \ref{Fig3} shows the visualization results of MPCount and L2HCount. We find that L2HCount's predicted result is more similar to GT's. Especially in high-density regions, L2HCount's predicted results are more reasonable, while MPCount tends to obtain overly dense prediction results. In low-density regions, L2HCount still obtains reasonable prediction results. The above analyses indicate that L2HCount adapts to crowd counting in low- and high-density regions.

\subsection{Comparison With State-of-the-Art (R$\to$A/Q)}

\begin{table}[!t]
\caption{Comparison with state-of-the-art methods (Training on the R dataset and further testing on A and Q datasets). “*” indicates that the model is first trained using the R dataset and then fine-tuned using a few samples from the A or Q datasets.
\label{tab:table7}}
\centering
\begin{tabular}{ccccccccccc}
\hline
\multicolumn{3}{c}{Source$\to${}Target} & \multicolumn{2}{c}{R$\to$A} & \multicolumn{2}{c}{R$\to${}Q} \\ \hline
Method                & DA         & DG         & MAE               & MSE               & MAE               & MSE               \\ \hline
ChfL \cite{ref48}                   & $\times$          & $\times$          & 234.0             & 347.1             & 330.2             & 564.4             \\
ChfL$^*$ \cite{ref48}                  & $\checkmark$          & $\times$          & \underline{142.5}             & \underline{242.9}             & \underline{309.3}             & \underline{518.4}             \\
MPCount \cite{ref14}               & $\times$          & $\checkmark$          & 178.0              & 265.7             & 337.1             & 522.9             \\
L2HCount(Ours)         & $\times$          & $\checkmark$          & \textbf{136.9}              & \textbf{235.8}             & \textbf{272.3}             & \textbf{450.3}             \\ \hline
\end{tabular}
\end{table}

To further validate our method's performance, we train our model on the R dataset and then test it directly on the A and Q datasets. Our method still achieves the best performance. On the A dataset, L2HCount achieves a 41.50$\%$ (136.9 vs. 234.0) improvement in MAE and a 32.07$\%$ (235.8 vs. 347.1) improvement in MSE compared to ChfL. On the Q dataset, L2HCount shows a 17.53$\%$ (272.3 vs. 330.2) improvement in MAE and a 20.22$\%$ (450.3 vs. 564.4) improvement in MSE. The performance improves significantly after fine-tuning the ChfL model with a few samples from the target dataset. On the A dataset, compared to the ChfL*, L2HCount shows a 3.93$\%$ (136.9 vs. 142.5) improvement in MAE and a 2.92$\%$ (235.8 vs. 242.9) improvement in MSE. On the Q dataset, L2HCount shows an 11.96$\%$ (272.3 vs. 309.3) improvement in MAE and a 13.14$\%$ (450.3 vs. 518.4) improvement in MSE.

\subsection{Ablation Study}

\begin{figure}[!t]
\centering
\includegraphics[width=3.5in]{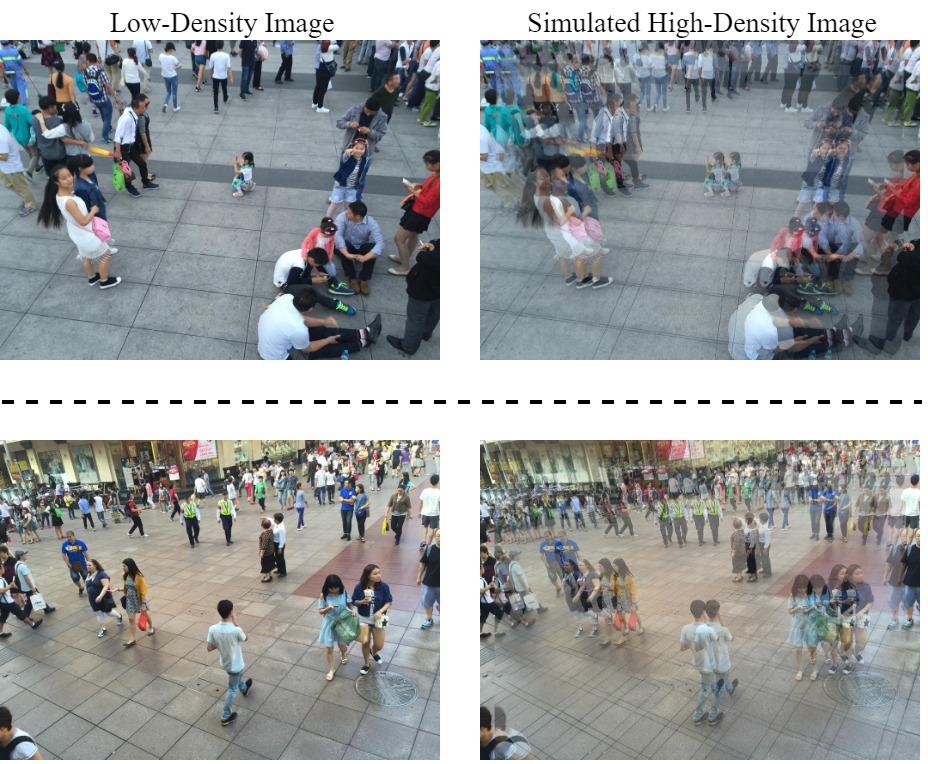}
\caption{Some samples of low- and simulated high-density images on the B dataset.}
\label{Fig4}
\end{figure}

\textit{1) The Impact of Training Data:} As shown in Fig. \ref{Fig4}, we generate simulated high-density images from low-density images for training. To ensure fairness in the training data, we retrain MPCount using both low- and simulated high-density images, labeling it as MPCount$\dag$. When tested on A dataset, L2HCount achieves a 19.87$\%$ (97.6 vs. 121.8) improvement in MAE  and a 14.23$\%$ (166.9 vs. 194.6) improvement in MSE compared to MPCount†. On the Q dataset, L2HCount shows a 6.12$\%$ (159.4 vs. 169.8) improvement in MAE and a 9.25$\%$ (267.7 vs. 295.0) improvement in MSE over MPCount†. This is because the simulated high-density image is blurry, making MPCount$\dag$ unable to effectively extract clear head features. However, the HFEM in L2HCount facilitates the learning of clear head features from simulated high-density images, enabling L2HCount to achieve superior performance.

\begin{table}[!t]
\caption{Ablation study for training data. To keep fairness, we simultaneously use low- and simulated high-density images as training data in MPCount$\dag$.\label{tab:table2}}
\centering
\begin{tabular}{ccccc}
\hline
Training data & \multicolumn{2}{c}{B$\to$A} & \multicolumn{2}{c}{B$\to$Q} \\ \hline
Method        & MAE               & MSE               & MAE               & MSE               \\ \hline
MPCount$\dag$       & 121.8                 & 194.6                 & 169.8                 & 295.0                 \\
L2HCount(Ours) & \textbf{97.6}                 & \textbf{166.9}                 & \textbf{159.4}                 & \textbf{267.7}                 \\ \hline
\end{tabular}
\end{table}

\textit{2) The Effect of Different Feature Scales in HFEM:} As shown in Table \ref{tab:table3}, the performance of our method has been significantly improved with the HFEM. Compared with the w/o HFEM, our method shows improvements of MAE by 20.84$\%$ (97.6 vs. 123.3) and MSE by 22.59$\%$ (166.9 vs. 215.6) on the A dataset. On the Q dataset, our method shows improvements of MSE by 25.16$\%$ (159.4 vs. 213.0) and MAE by 28.56$\%$ (267.7 vs. 374.7). We find that using the HFEM on the smaller-scale feature map can achieve better performance. We also attempt to use HFEM simultaneously on S3 and other scales, but the performance does not improve further.

Based on the above analysis, we draw the following conclusions: First, the HFEM guides the encoder's feature learning in the high-density branch, enabling it to learn clear feature maps. Second, the HFEM is more effective for smaller-scale feature maps. The smaller-scale feature map retains more critical semantic information, which enhances HFEM's ability to provide effective feature learning guidance.

\begin{figure}[!t]
\centering
\includegraphics[width=3.5in]{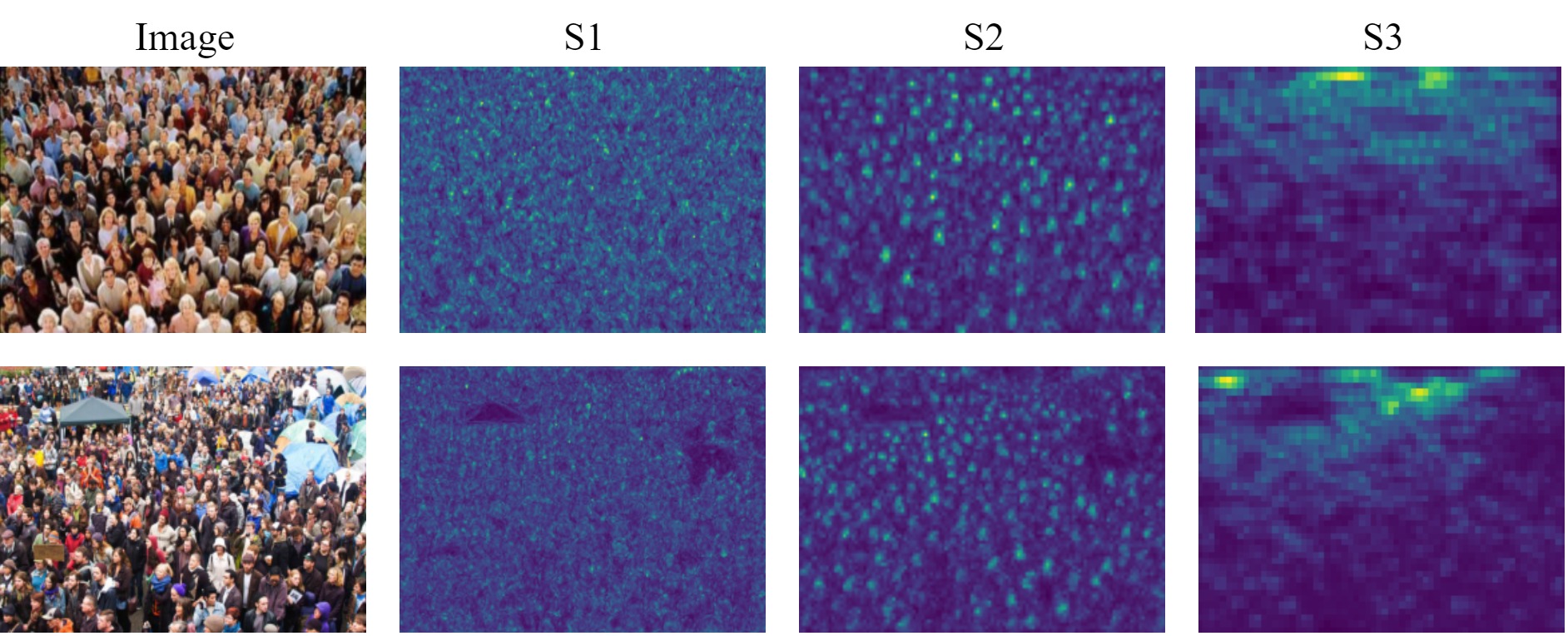}
\caption{Some samples of feature maps are extracted by the encoder on each scale.}
\label{Fig5}
\end{figure}

As shown in Fig. \ref{Fig5}, we visualize feature maps of different scales. Large-scale feature maps focus on texture and edge information, while small-scale feature maps focus on crowd distribution information. Since crowd distribution information is crucial for crowd counting, using HFEM at the S3 scale can lead to better performance.

\begin{table}[!t]
\caption{Ablation study for different feature scales in HFEM. S1, S2, and S3 represent three different scales of feature maps extracted by the encoder, with the size of the feature maps decreasing in sequence.\label{tab:table3}}
\centering
\begin{tabular}{ccccccc}
\hline
\multicolumn{3}{c}{HFEM}                                                                                                                               & \multicolumn{2}{c}{B$\to$A} & \multicolumn{2}{c}{B$\to$Q} \\ \hline
\begin{tabular}[c]{@{}c@{}}S1\end{tabular} & \begin{tabular}[c]{@{}c@{}}S2\end{tabular} & \begin{tabular}[c]{@{}c@{}}S3\end{tabular} & MAE               & MSE               & MAE               & MSE               \\ \hline
                                                  &                                                    &                                                    & 123.3                 & 215.6                 & 213.0                 & 374.7                 \\
$\checkmark$                                                  &                                                    &                                                    & 116.9                 & 201.1                 & 191.9                 & 335.2                 \\
                                                   & $\checkmark$                                                  &                                                    & 113.9                 & 194.9                 & 182.7                 & 326.8                 \\
                                                   &                                                    & $\checkmark$                                                  & \textbf{97.6}                 & \textbf{166.9}                 & \textbf{159.4}                 & \textbf{267.7}                 \\
                                                   & $\checkmark$                                                  & $\checkmark$                                                  & 101.0                 & 178.3                 & 180.4                 & 316.7                 \\
$\checkmark$                                                  &                                                    & $\checkmark$                                                  & 123.2                 & 197.2                 & 212.6                 & 382.5                 \\ \hline
\end{tabular}
\end{table}

\begin{figure}[!t]
\centering
\includegraphics[width=3.5in]{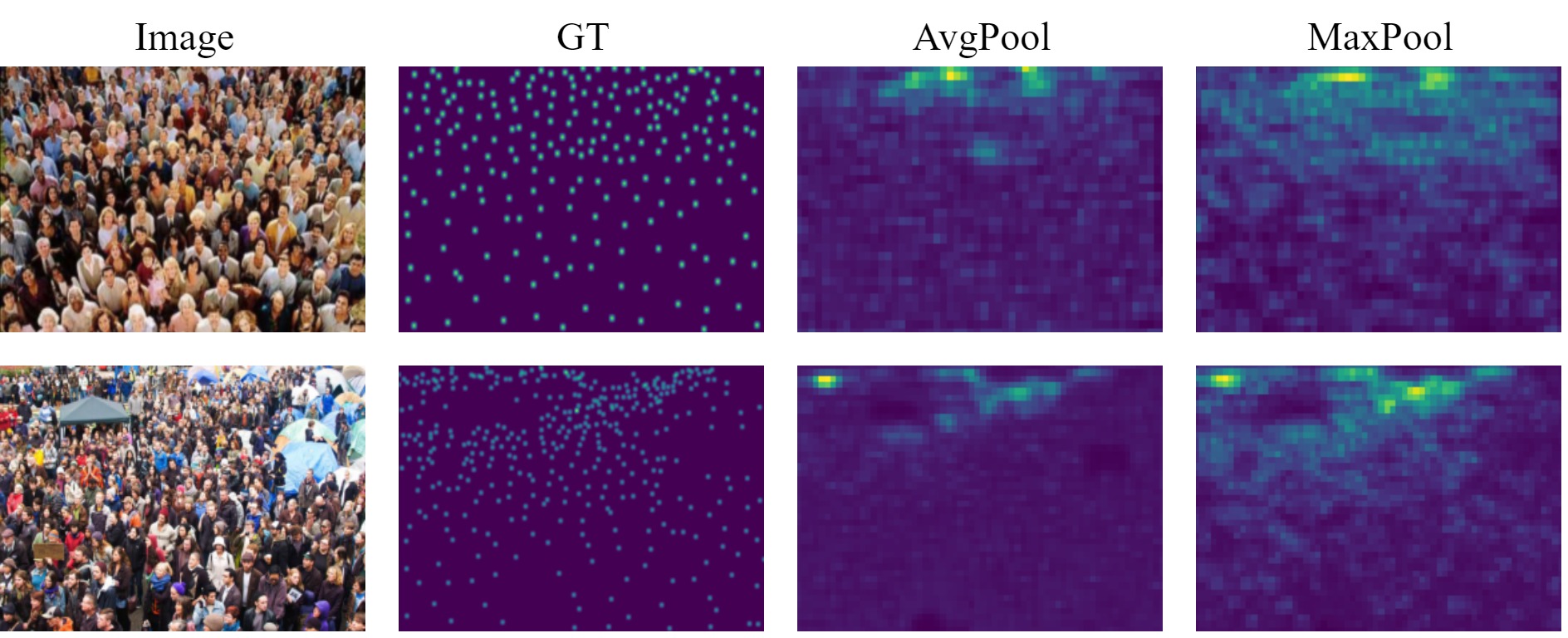}
\caption{The feature maps are obtained by average pooling and max pooling based on the feature maps of the S3 scale.}
\label{Fig6}
\end{figure}

\textit{3) The Effect of Pooling Strategy in HFEM:} We explore the effects of average pooling, max pooling, and w/o pooling on HFEM. As shown in Table \ref{tab:table4}, using max pooling to process feature maps achieves the best performance. Compared to w/o pooling, our method improves the MAE performance by 13.48$\%$ (97.6 vs. 112.8) and the MSE performance by 9.49$\%$ (166.9 vs. 184.4) on the A dataset. On the Q dataset, our method improves MAE performance by 5.90$\%$ (159.4 vs. 169.4) and MSE performance by 7.02$\%$ (267.7 vs. 287.9). We also attempt to use average and max pooling together to process feature maps, but this does not yield further improvements. As shown in Fig. \ref{Fig6}, max pooling effectively preserves significant activations on feature maps, making it the best choice for HFEM.

\begin{table}[!t]
\caption{Ablation study for pooling strategy in HFEM.\label{tab:table4}}
\centering
\begin{tabular}{cccccc}
\hline
\multicolumn{2}{c}{\begin{tabular}[c]{@{}c@{}}HFEM\end{tabular}} & \multicolumn{2}{c}{B$\to$A} & \multicolumn{2}{c}{B$\to$Q} \\ \hline
AvgPool                                & MaxPool                               & MAE               & MSE               & MAE               & MSE               \\ \hline
                                      &                                       & 112.8                 & 184.4                 & 169.4                 & 287.9                 \\
$\checkmark$                                      &                                       & 126.2                 & 208.1                 & 214.0                 & 351.4                 \\
                                       & $\checkmark$                                     & \textbf{97.6}                 & \textbf{166.9}                 & \textbf{159.4}                 & \textbf{267.7}                 \\
$\checkmark$                                      & $\checkmark$                                     & 103.6                 & 172.3                 & 199.4                 & 355.2                 \\ \hline
\end{tabular}
\end{table}

\textit{4) The Effect of LDCM and HDCM in DDMEM:} As shown in Table \ref{tab:table5}, the best performance can be achieved by using LDCM and HDCM simultaneously for feature map re-encoding. On the A dataset, our method improves MAE by 1.01$\%$ (97.6 vs. 98.6) and MSE by 4.25$\%$ (166.9 vs. 174.3) compared to only using LDCM. On the Q dataset, our method enhances MAE by 10.45$\%$ (159.4 vs. 178.0) and MSE by 14.47$\%$ (267.7 vs. 313.0) over only using LDCM.

\begin{figure}[!t]
\centering
\includegraphics[width=3.5in]{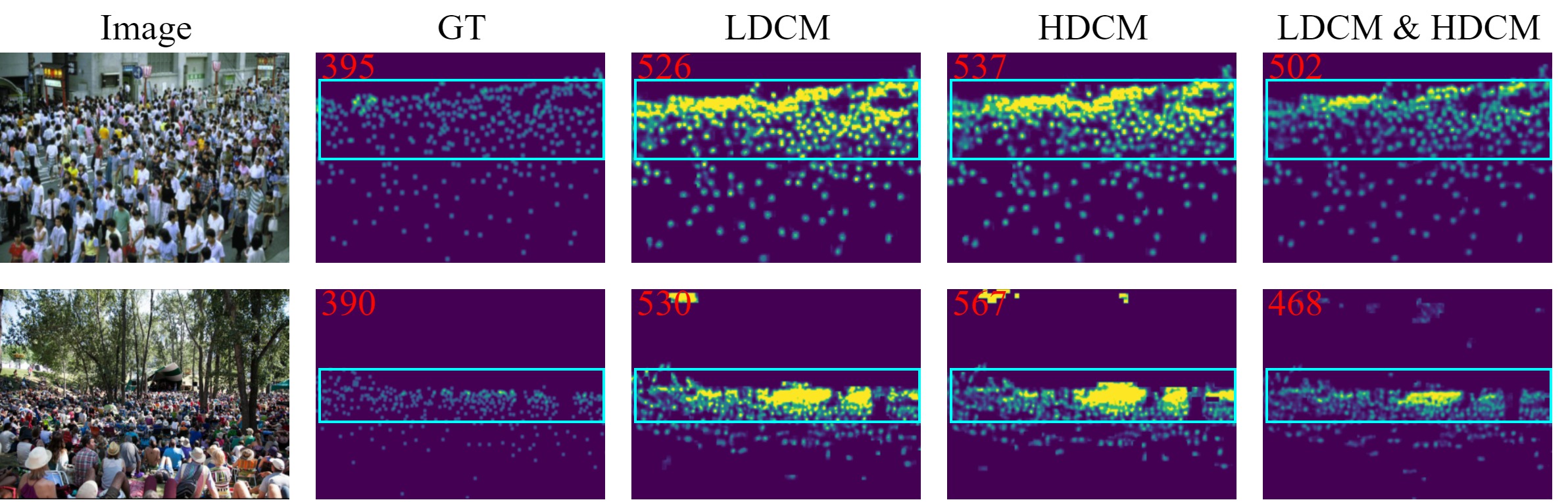}
\caption{Visualization of the predicted results. We show the predicted results using different crowd memory, including LDCM, HDCM, and LDCM \& HDCM. LDCM $\&$ HDCM predicts more reasonable results in the rectangle regions.}
\label{Fig7}
\end{figure}

As shown in Fig. \ref{Fig7}, the density maps predicted by LDCM \& HDCM are more similar to GT. Since LDCM and HDCM learn different crowd distribution patterns, using them simultaneously provides richer information for crowd counting.

\begin{table}[!t]
\caption{Ablation study for the effect of LDCM and HDCM in DDMEM.\label{tab:table5}}
\centering
\begin{tabular}{cccccc}
\hline
\multicolumn{2}{c}{DDMEM}                                                                            & \multicolumn{2}{c}{B$\to$A} & \multicolumn{2}{c}{B$\to$Q} \\ \hline
\begin{tabular}[c]{@{}c@{}}LDCM\end{tabular} & \begin{tabular}[c]{@{}c@{}}HDCM\end{tabular} & MAE               & MSE               & MAE               & MSE               \\ \hline
$\checkmark$                                                   & \multicolumn{1}{l}{}                                & 98.6                 & 174.3                 & 178.0                 & 313.0                 \\
\multicolumn{1}{l}{}                                & $\checkmark$                                                   & 101.4                 & 174.5                 & 179.4                 & 315.6                 \\
$\checkmark$                                                   & $\checkmark$                                                   & \textbf{97.6}                 & \textbf{166.9}                 & \textbf{159.4}                 & \textbf{267.7}                 \\ \hline
\end{tabular}
\end{table}

\textit{5) The Effect of Fusion Strategy in DDMEM:} As shown in Eq. \ref{eq4}, this paper adopts the concatenation strategy to fuse the features re-encoded by LDCM and HDCM. In addition, we explore other fusion strategies, such as addition and adaptive selection.
\begin{itemize}
 \item Addition strategy: We directly perform element-wise addition on the features re-encoded by LDCM and HDCM.
 \item Adaptive selection strategy: We use re-encoded features by LDCM and HDCM to generate corresponding density maps, respectively. Then, we apply the adaptive module proposed in \cite{ref5} to adaptively select the proper prediction manner from the two density maps to generate the final density map.
\end{itemize}

As shown in Table \ref{tab:table6}, we achieve the best performance with the concatenation strategy. Our method significantly outperforms the addition strategy, achieving a 19.41$\%$ (97.6 vs. 121.1) improvement in MAE and a 22.34$\%$ (166.9 vs. 214.9) improvement in MSE on the A dataset. On the Q dataset, our method enhances MAE by 17.19$\%$ (159.4 vs. 192.5) and MSE by 20.73$\%$ (267.7 vs. 337.7). When compared to the adaptive selection strategy, our method shows a 3.08$\%$ improvement in MAE (97.6 vs. 100.7) and a 5.81$\%$ improvement in MSE (166.9 vs. 177.2) on the A dataset. On the Q dataset, our method achieves a 12.08$\%$ (159.4 vs. 181.3) improvement in MAE and a 16.76$\%$ (267.7 vs. 321.6) improvement in MSE.

\begin{figure}[!t]
\centering
\includegraphics[width=3.5in]{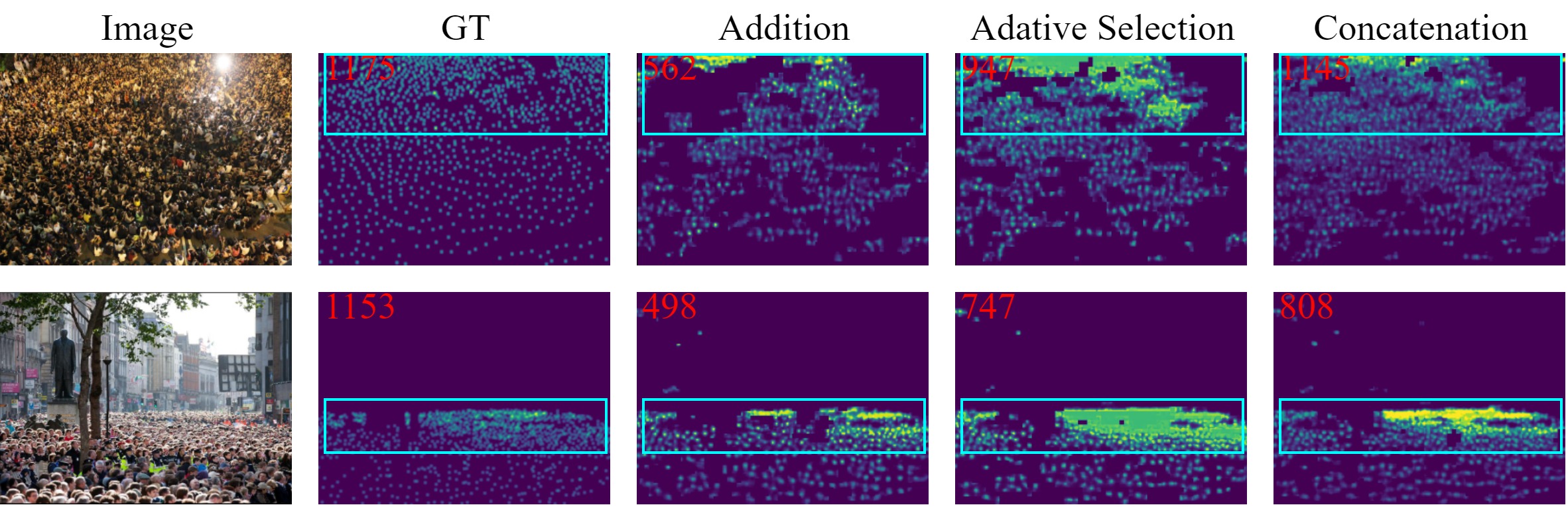}
\caption{Visualization of the predicted results. We show the predicted results using different fusion strategies, including addition, adaptive selection, and concatenation. The concatenation strategy predicts more reasonable results in the rectangle regions.}
\label{Fig8}
\end{figure}

As shown in Fig. \ref{Fig8}, the density map predicted by our method is more reasonable, especially in high-density regions. Based on the above analysis, the concatenation strategy effectively preserves both low-density and high-density patterns, enabling more accurate density map prediction for crowd counting.

\begin{table}[!t]
\caption{Ablation study for the fusion strategy of LDCM and HDCM in DDMEM.\label{tab:table6}}
\centering
\begin{tabular}{cllcccc}
\hline
\multicolumn{3}{c}{DDMEM}              & \multicolumn{2}{c}{B$\to$A} & \multicolumn{2}{c}{B$\to$Q} \\ \hline
\multicolumn{3}{c}{Fusion Strategy}    & MAE               & MSE               & MAE               & MSE               \\ \hline
\multicolumn{3}{c}{Addition}           & 121.1                 & 214.9                 & 192.5                 & 337.7                 \\
\multicolumn{3}{c}{Adaptive Selection} & 100.7                 & 177.2                 & 181.3                 & 321.6                 \\
\multicolumn{3}{c}{Concatenation}      & \textbf{97.6}                 & \textbf{166.9}                 & \textbf{159.4}                 & \textbf{267.7}                 \\ \hline
\end{tabular}
\end{table}

\section{Conclusion}

In this paper, we propose L2HCount, a framework designed to generalize the model from the low- to high-density scene. To generalize our model well on the high-density scene, we design HDSM to generate simulated high-density images based on low-density images, enabling our model to learn the pattern of high-density images. In addition, we propose GTGM to generate the corresponding ground-truth crowd annotations for simulated high-density images. However, simulated images introduce two new problems: image blurring and loss of low-density image characteristics. To address the first problem, we design HFEM to guide the model in extracting clear head features from simulated high-density images. To address the second problem, we propose DDMEM, which employs LDCM and HDCM to learn the patterns of low- and simulated high-density scenes, respectively. Finally, we conduct extensive experiments on STB, RGBT-CC, STA, and QNRF datasets to demonstrate the effectiveness of our method.

\section*{Acknowledgments}
This work was supported by the National Natural Science Foundation of China (Grant No. 62173045), the Beijing Natural Science Foundation under Grant F2024203115, and the China Postdoctoral Science Foundation under Grant Number 2024M750255.

\end{document}